\documentclass[runningheads]{llncs}
\usepackage[T1]{fontenc}
\usepackage{xcolor}
\usepackage{graphicx}

\usepackage[T1]{fontenc}
\usepackage{graphicx}
\usepackage{cite}
\usepackage{amsmath,amssymb,amsfonts}
\usepackage{algorithmic}

\usepackage{textcomp}
\usepackage{xcolor}

\usepackage{algorithm}
\usepackage{color}

\usepackage{caption}

\usepackage{subfigure}

\usepackage{marvosym}

\begin{document}
\title{Mitigating the Impact of Noisy Edges on Graph-Based Algorithms via Adversarial Robustness Evaluation}
\titlerunning{Mitigating the Impact of Noisy Edges}
%
\author{Yongyu Wang* \and
Xiaotian Zhuang*}
\authorrunning{Y. Wang et al.}
%
\institute{JD Logistics, Beijing 101111, China\\
}

\maketitle              
\renewcommand{\thefootnote}{}
\footnotetext{*These authors contributed equally and are co-first authors: Yongyu Wang and Xiaotian Zhuang}

\renewcommand{\thefootnote}{}
\footnotetext{Correspondence to: wangyongyu1@jd.com}

\begin{abstract}
Given that no existing graph construction method can generate a perfect graph for a given dataset, graph-based algorithms are often affected by redundant and erroneous edges present within the constructed graphs. In this paper, we view these noisy edges as adversarial attack and propose to use a spectral adversarial robustness evaluation method to mitigate the impact of noisy edges on the performance of graph-based algorithms. Our method identifies the points that are less vulnerable to noisy edges and leverages only these robust points to perform graph-based algorithms. Our experiments demonstrate that our methodology is highly effective and outperforms state-of-the-art denoising methods by a large margin.

\end{abstract}
\section{Introduction}
\label{sec:intro}

For many graph-based algorithms, the initial phase entails the construction of a graph from the provided dataset \cite{von2007tutorial}. This graph is structured such that each node corresponds to an individual data point, while the edges delineate the interrelations among these points. Given the intrinsic uniqueness of each dataset, it is impractical to expect a universal graph construction method that can precisely cater to all datasets. Consequently, the graphs generated may contain a significant number of erroneous and superfluous edges, commonly referred to as noisy edges. These noisy edges can profoundly degrade the performance of graph-based algorithms \cite{premachandran2013consensus,Wang_2022_BMVC}.

Currently, the most widely used method for graph construction is the k-nearest neighbor (k-NN) graph. In a k-NN graph, each node is connected to its k nearest neighbors. This method possesses a strong capability to capture the local manifold \cite{roweis2000nonlinear}, which is why it has become the predominant graph construction technique for the majority of graph-based algorithms. However, k-NN graph has a tendency to include noisy edges \cite{yang2012affinity}. Within datasets, the distribution and characteristics of points are neither uniform nor consistent, rendering the use of a uniform k value for all points imprecise \cite{premachandran2013consensus}. Furthermore, the metric used to measure the distance between two points is also problematic. Whether it is Euclidean distance, cosine similarity, or any other distance metric, each has its limitations \cite{sarwar2001item}. Consequently, the distances measured are not always accurate. Therefore, the k nearest neighbors in a k-NN graph may not necessarily be the ones that should be connected.

Addressing this issue is extremely challenging, primarily because, for the task of capturing the underlying structure of a dataset using a graph, there is no ground truth solution. Thus, only heuristic methods are available to remove noisy edges from the graph. Among the most representative works in this area is the consensus method proposed by \cite{premachandran2013consensus}, which extracts consensus information from a given k-NN graph. In this method, edges with a consensus value below a certain threshold are pruned. However, this approach, while removing noise, also eliminates a substantial number of non-noisy, useful edges. \cite{Wang_2022_BMVC} proposed a spectral framework to detect non-critical, misleading, and superfluous edges in the graph . However, the gains in algorithmic solution quality are still relatively modest.

This paper introduces a novel method aimed at enhancing the noise resilience of graph-based algorithms. Unlike approaches that attempt to remove noisy edges from the graph, our method employs a spectral adversarial robustness evaluation method to identify a small amount of robust nodes that exhibit strong resistance to noise. We then utilize only these robust nodes to complete the graph analysis tasks. Our approach not only enhances the solution quality of graph-based algorithms but can also help to reduce the computational cost of these algorithms. The majority of graph-based algorithms have a time complexity of at least \textbf{$O(n^2)$}, and many are \textbf{$O(n^3)$}, such as spectral clustering algorithms \cite{von2007tutorial}, where $n$ is the number of nodes in the graph. Unlike traditional approaches that involve all nodes, our method only requires the robust nodes, thus substantially lowering the computational cost.

The main contributions of this work are as follows:

\begin{enumerate}
  \item We view noisy edges in graphs as adversarial attacks and propose to use a spectral proactive defense approach to fundamentally address this issue. \\
  
  \item In contrast to existing approaches that address the issue of noisy edges from the perspective of the edges themselves, our method approaches the problem from the node perspective. Recognizing the inherent challenge in discerning whether a specific edge in a graph is noise or necessary, we propose a solution that focuses on identifying nodes that are not vulnerable to noisy edges. By doing so, we aim to reconstruct a robust dataset that is resilient to the presence of noisy edges.\\
  
  \item We show that by utilizing only a small number of robust nodes, significant improvements can be achieved in both the accuracy and efficiency for graph-based algorithms.\\
  
\end{enumerate}

\section{Preliminaries}

\subsection{Adversarial attack and defense}

Adversarial attack aims to misled machine learning models by providing deceptive inputs, such as samples with intentional disturbances \cite{goodfellow2018making,fawzi2018analysis}, which are commonly known as
adversarial examples. \cite{szegedy2013intriguing,nguyen2015deep, moosavi2016deepfool} have demonstrated that machine learning models are often highly vulnerable to adversarial attacks.

To defend against adversarial attacks, many methods have been proposed. These methods can be categorized into two types: reactive defenses and proactive defenses. Reactive defenses concentrate on detecting adversarial examples within the model's inputs, as investigated by \cite{feinman2017detecting,metzen2017detecting,yang2020ml}. In contrast, proactive defenses seek to bolster the robustness of the models, making them less vulnerable to the influence of adversarial examples, such as the methods presented in \cite{agarwal2019improving,liu2018adv}.

\subsection{Spectral clustering}\label{distortionPerspective}
Spectral clustering is one of the most representative and widely applied graph-based algorithms. It can often outperform traditional clustering algorithms, such as k-means algorithms, due to its ability to extract structural features of the dataset from the graph representation \cite{von2007tutorial}. There are three common spectral clustering algorithms used in practice, i.e., unnormalized spectral clustering \cite{von2007tutorial} and two normalized spectral clustering methods \cite{shi2000normalized,ng2001spectral}. These algorithms are quite similar, apart from using different graph Laplacians. As shown in Algorithm \ref{alg:sc}, typical spectral clustering algorithms can be divided into three steps: 1) construct a data graph according to the entire data set, 2) embed all data points into $k$-dimensional space using eigenvectors of $k$ bottom nonzero eigenvalues of the graph Laplacian, and 3) perform k-means algorithm to partition the embedded data points into $k$ clusters.

\begin{algorithm}[!htbp]
\small { \caption{Unnormalized Spectral Clustering Algorithm} \label{alg:sc}
\textbf{Input:} A data set $D$ with $N$ samples $x_1,...,x_N \in {R}^{d}$, number of clusters $k$.\\
\textbf{Output:} Clusters $C_1$,...,$C_k$.\\
\begin{algorithmic}[1]
    \STATE Construct a graph $G$ from the input data ; \\
    \STATE Compute the adjacency matrix $A_G$, and diagonal matrix $D_G$ of graph $G$; \\
    \STATE Obtain the unnormalized Laplacian matrix $L_G$=$D_G$-$A_G$;\\
    \STATE Compute the eigenvectors $u_1$,...$u_k$ that correspond to the bottom k nonzero eigenvalues of $L_G$;\\
    \STATE Construct $U \in \mathbb{R}^{n \times k}$, with $k$ eigenvectors of $L_G$ stored as columns;\\
    \STATE Perform k-means algorithm to partition the rows of $U$ into $k$ clusters and return the result.\\
\end{algorithmic}
}
\end{algorithm}

\section{Method}

\subsection{Examining Graph-Based Machine Learning Models from the Perspective of Adversarial Attacks}

Machine learning models are fundamentally mechanisms that map inputs to outputs via feature transformation. For instance, deep neural networks distill the original feature vectors of data through successive layers. Similarly, algorithms like Support Vector Machine (SVM) and Support Vector Clustering (SVC) utilize kernel functions to map the original feature space of data into a higher-dimensional feature space \cite{ben2001support}. Adversarial attacks on machine learning models aim to mislead this mapping process. 

In graph-based machine learning algorithms, the graph plays a pivotal role as the algorithm extracts structural information from the graph to transform input data, thereby mapping the inputs to outputs. Therefore, from the perspective of adversarial attacks, perturbing the graph can disrupt the mapping process of the graph-based machine learning model.

Suppose there exists a ground-truth 'perfect' graph for a graph-based algorithm. In that case, any discrepancies between the actual graph constructed by our graph-building algorithm and the ground-truth graph could be considered as adversarial attacks applied to the ground-truth graph. Therefore, we propose to use adversarial defense methods to manage these noisy edges.

\subsection{A Proactive Defense Strategy from the Node Perspective to Mitigate the Impact of Noisy Edges}

Given that the ideal connectivity of edges within a graph is perpetually unknown, previous research focused on managing noisy edges directly from the edge perspective has yielded limited success \cite{premachandran2013consensus,Wang_2022_BMVC}. In this paper, we propose to address the issue of noisy edges from the perspective of nodes.

Graph-based machine learning models, much like other machine learning models, predominantly focus on tasks associated with data points, such as classifying these points or predicting values associated with them. Graph-based algorithms distinguish themselves by utilizing the relationships signified by edges to assist in accomplishing tasks that are centered on the nodes. Edges in a graph play a twofold role. While they contain significant structural information that can aid in achieving more accurate data transformations—such as those utilized by graph neural networks to harness the associative information between nodes for improved performance—they can also be detrimental. A multitude of incorrect or superfluous edges can indeed degrade the performance of machine learning models. By enhancing the nodes' resilience to noisy edges, our method can exploit the beneficial edges to boost algorithmic performance without being adversely affected by the harmful ones. In this paper, we propose enhancing the nodes' resilience to noisy edges as a means to achieve improved algorithm performance by utilizing beneficial edges while concurrently mitigating the negative impact of harmful edges, when both coexist within the graph.

\subsection{Adversarial Robustness Evaluation}

In order to identify the nodes with strong resilience to noisy edges, we first evaluate the robustness of data points to noisy edges. 

Inspired by \cite{weng2018evaluating}, \cite{cheng2021spade} proposed that the adversarial robustness of a given machine learning model can be measured by examining the distortion between the manifolds of the input feature space and the output feature space, by leveraging the generalized Courant-Fischer theorem \cite{spielman2012spectral}. In this section, we employ this method to evaluate the robustness of each data point in the spectral clustering model. The specific steps are as follows:\\

\begin{itemize}
    \item { Given data set $D$ with $N$ samples $x_1,...,x_N \in {R}^{d}$ and its number of clusters $k$, we first construct a k-NN graph $G_{\text{input}}$ to capture the data manifold in the original $d$-dimensional feature space.}\\
    \item {We perform the spectral embedding step in the spectral clustering algorithm to map the data points from the original $d$-dimensional space into $k$-dimensional spectral space to obtain data set $U$ with the points in the embedded feature space.}\\
    \item {We construct a k-NN graph $G_{\text{output}}$ to capture the data manifold in the embedded $k$-dimensional feature space.}\\
    
    \item {Based on the generalized Courant-Fischer theorem \cite{spielman2012spectral}, \cite{cheng2021spade} has further shown that the generalized eigenpairs of $L^+_{output} L_{input}$ can be used to estimate the robustness of each point, where  $L^+_{output}$ denotes the Moore–Penrose pseudoinverse of the graph Laplacian of $G_{\text{output}}$ and $L_{input}$ denotes the graph Laplacian matrix of $G_{\text{input}}$. To this end, we construct the following eigensubspace matrix $\mathbf{V}_k \in \mathbb{R}^{N \times k}$:
    \begin{equation}
\mathbf{V}_k {=} \left[ \mathbf{v}_1 \sqrt{\lambda_1}, \ldots, \mathbf{v}_k \sqrt{\lambda_k} \right],
\end{equation}
where $\lambda_1, \lambda_2, \ldots, \lambda_k$ represent the first $k$ largest eigenvalues of $L^+_{output} L_{input}$ and $\mathbf{v}_1, \mathbf{v}_2, \ldots, \mathbf{v}_k$ are the corresponding eigenvectors.}\\

\item {Finally, a metric called spade score for evaluating the adversarial robustness of a specific node $i$ can be calculated as follows \cite{cheng2021spade}:

\begin{equation}
\text{spade}(i) = \frac{1}{|\mathcal{N}_(i)|} \sum_{j \in \mathcal{N}_(i)} \| \mathbf{V}_k^\top \mathbf{e}_{i,j} \|_2^2;
\end{equation}
where $j \in \mathcal{N}_(i)$ denotes the $j$-th neighbor of node $i$ in graph $G_{\text{input}}$, and $\mathcal{N}_(i) \subseteq V$ denotes the node set including all the neighbors of node $i$, \( \mathbf{e}_{i,j} = \mathbf{e}_{i} - \mathbf{e}_{j} \), and \( \mathbf{e}_{o} \in \mathbb{R}^{N} \) denotes the standard basis vector with the \( i \)-th element being 1 and others being 0. A larger $\text{spade}(i)$ implies that node $i$ is likely more vulnerable to adversarial attacks. 

}

\end{itemize}

The above procedures are efficient: constructing k-nearest graph can be done within $O(|n|log|n|)$ time \cite{malkov2018efficient}; The spade score can be computed in nearly-linear time leveraging recent fast Laplacian solvers \cite{kyng2016approximate}.

\subsection{A Multi-Level Algorithm Framework Based on Robust Node Set}
We calculate the spade score for all data points and sort them in ascending order. We then select a small number of data points with the lowest spade score, which correspond to the highest robustness, to form a robust subset. 
After obtaining the robust node set, we perform spectral clustering exclusively on this set to group the robust nodes into $k$ clusters. For each cluster, we calculate its centroid. Subsequently, each non-robust data point is assigned to the cluster whose centroid is closest to it.

\section{Experiment}

In this section, we apply the proposed method to k-NN graph and use unnormalized spectral clustering to demonstrate its effectiveness. We assess the efficacy of our proposed approach by evaluating its ability to improve solution quality, as well as its capacity to increase the operational efficiency of the algorithm.

\subsection{Data sets}

Experiments are performed using the following two real-world benchmark data sets:

\begin{itemize}
    \item {\textbf{USPS:} includes   $9,298$ images of USPS hand written digits with $256$ attributes.}\\
    \item {\textbf{MNIST:} the machine learning field's most recognized benchmark, features 60,000 training and 10,000 test images of handwritten digits, each with 784 attributes. We evaluate our methods using its test set.}
\end{itemize}

\subsection{Metric}

To assess the solution quality of spectral clustering, we use the accuracy metric. It is defined as:\begin{equation}\label{eqn:scale}
ACC= \frac{\sum\limits_{j = 1}^n  {\delta {(y_i,map(c_i))}}}{{n}},
\end{equation}

where $n$ represents the total count of data instances within the dataset, $y_i$ denotes the ground-truth label as provided by the dataset, and $C_i$ signifies the label ascribed by the clustering algorithm. The function $\delta (x,y)$ is a delta function, stipulated as: $\delta (x,y)$=1 for $x=y$, and $\delta (x,y)$=0,  otherwise. The function $map(\bullet)$  serves as a permutation mapping that correlates each cluster index $c_i$ with an equivalent ground truth label, a process which can be efficiently accomplished utilizing the Hungarian algorithm \cite{papadimitriou1998combinatorial}. An elevated $ACC$ value is indicative of superior clustering performance.\\

\subsection{Compared Algorithms}

We compare our method against both the baseline and the state-of-the-art techniques for handling noisy edges in graphs. The specifics are as follows:\\

\begin{itemize}
    \item {\textbf{k-nearest neighbor graph:} For the value of k in the k-NN graph for the USPS and the MNIST data sets, we use the setting in \cite{szlam2009total,Wang_2022_BMVC}: k is set to 10; }\\
    \item {\textbf{Consensus method:} the state-of-the-art technique for selecting neighborhoods to construct affinity graphs. This method strengthens the graph's robustness by incorporating consensus information from various neighborhoods in a specified kNN graph \cite{premachandran2013consensus};}\\
    \item {\textbf{Spectral edge sparsification method:} The state-of-the-art method for detecting non-critical, misleading, and superfluous edges in the graph \cite{Wang_2022_BMVC}.}\\
\end{itemize}

\vspace{-0.1in}

\subsection{Results Of Solution Quality}

Table~\ref{table:compare3} shows the solution quality of graph-based spectral clustering algorithm on the USPS and the MNIST data sets.

\begin{table}[!htbp]
\centering
\caption{Clustering Accuracy (\%)}\label{table:compare3}
\scalebox{1.2}{ 
\setlength{\tabcolsep}{12pt} 
\begin{tabular}{ c c c c c }
\hline
Data Set & $k$-NN & Consensus & Spectral Spar & Ours \\
\hline
USPS & 64.31 & 68.54 & 70.74 & 78.87 \\
MNIST & 59.68 & 61.09 & 60.09 & 70.40 \\
\hline
\end{tabular}}
\end{table}

The clustering outcomes of our approach are derived by selecting the top 2,000 and 1,500 nodes with the highest robustness from the USPS and MNIST data sets, respectively. It is evident that our method surpasses the baseline kNN graph by over 14\% and 10\% in accuracy on the USPS and MNIST data sets, respectively, demonstrating the effectiveness of our approach in improving the solution quality. Furthermore, the results achieved by our approach, which surpasses the second-best denoising method by margins of 8\% for the USPS data set and 9\% for the MNIST data set, validate the advantage of employing the algorithm on nodes with reduced sensitivity to noisy edges over existing methods that concentrate on the elimination of such edges. 

It can be seen that methods aimed at resolving noisy issues by removing noisy edges do manifest a clear improvement on the USPS data set, although they fall short of the enhancements our method provides when compared to the baseline. On the MNIST data set, however, their effects are marginal, with the consensus method and spectral sparsification method achieving only 1.5\% and 0.5\% increases in accuracy, respectively. We conjecture that this is attributable to the MNIST data set containing a greater number of features than the USPS data set, which may include more non-robust features that can induce noisy edges. Consequently, the task of excising noisy edges from MNIST is significantly more formidable. Existing denoising methods face difficulties in purging noisy edges without also eliminating beneficial edges. In contrast, our approach, which involves selecting nodes that are robust to noise and executing graph-based algorithms exclusively among these robust nodes, serves as an attack-agnostic method. This renders it highly effective for both the USPS and MNIST data sets.

\subsection{Efficacy of Resolving Computational Bottleneck}

In spectral clustering algorithm, the first step of constructing the kNN graph can be completed within a time complexity of $O(|n|log|n|)$ \cite{malkov2018efficient}, where $n$ is number of nodes in the graph. The third step, $k$-means, can also be performed in linear time \cite{pakhira2014linear}. However, the second step, eigen-decomposition, has a time complexity of \textbf{$O(n^3)$}, making it the computational bottleneck of the entire algorithm, as well as the dominant term in the time complexity analysis of the algorithm.

In our method, eigen-decomposition is only performed for a subset composed of a small number of robust nodes, thereby significantly reducing the computational cost.

\begin{figure}[!htbp]
\centering\includegraphics[scale=0.25]{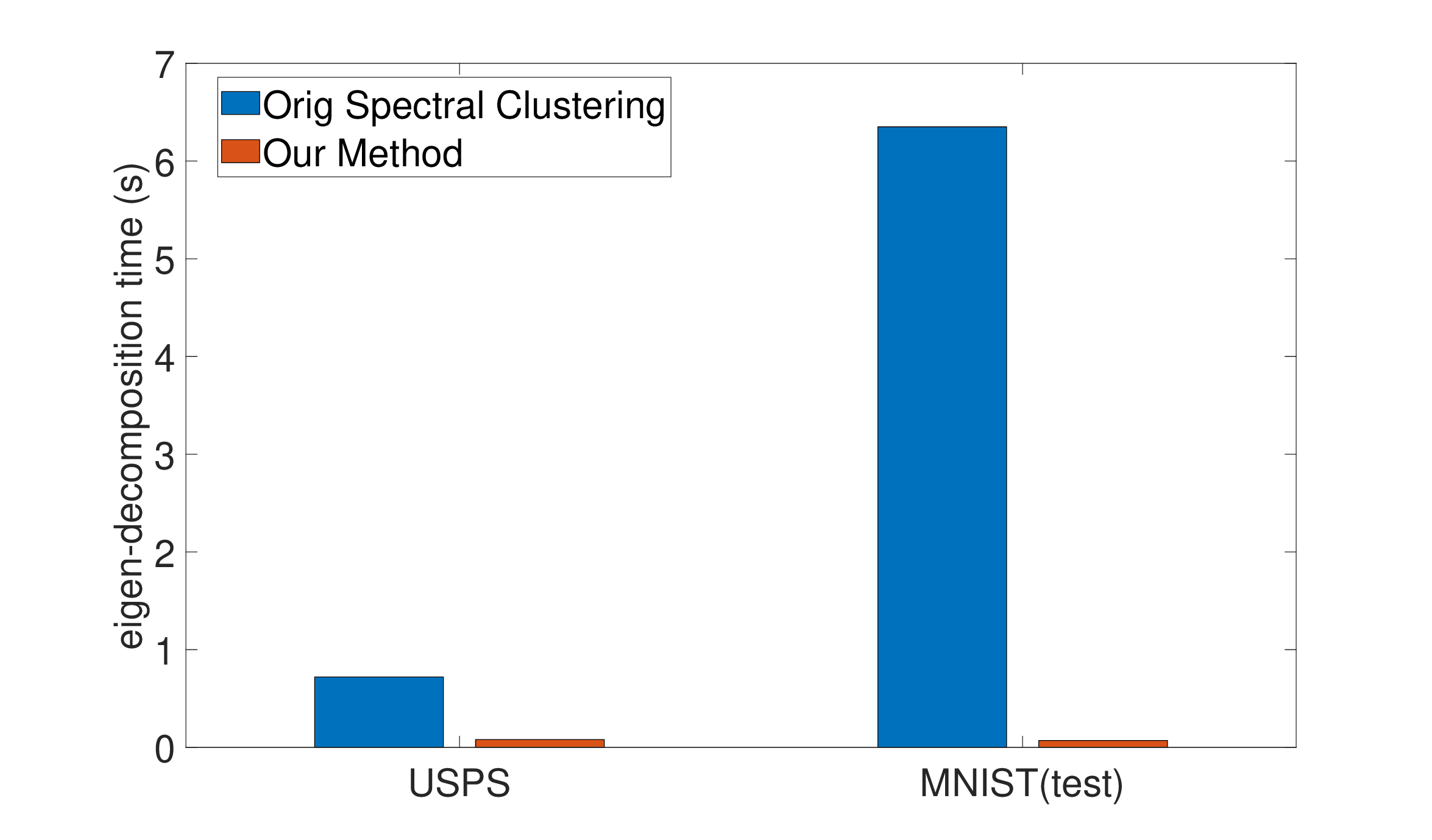}
\caption{Eigen-decomposition time of original node set and robust node set.\protect\label{fig:eigentime.eps}}
\end{figure}

Figure \ref{fig:eigentime.eps} and Table~\ref{table:density} show the eigen-decomposition time of the original full data set and our selected robust node set. It can be observed that by utilizing the selected robust node set, eigen-decomposition has been accelerated by a factor of 9 for the USPS data set and 90 for the MNIST data set, respectively. It is expected that the proposed method will be a key enabler for running computationally expensive graph-based algorithms in scenarios that require extremely fast response times and on devices with limited computational capabilities.

\begin{table}[!htbp]
\centering 
\caption{Eigen-decomposition Time (s)}
\label{table:compare4}
\scalebox{1.2}{ 
\setlength{\tabcolsep}{12pt} 
\begin{tabular}{ c c c }
\hline
Data Set & Original Node Set & Robust Node Set \\
\hline
USPS & 0.72 & 0.08  \\
MNIST(test) & 6.35 & 0.07  \\
\hline
\end{tabular}}\label{table:density}
\end{table}

\subsection{Parameter Discussion}
In our method, the construction of $G_{\text{input}}$ and  $G_{\text{output}}$ involves the selection of k in the k-NN algorithm. Our preliminary experiments indicate that when k is large enough to adequately capture the underlying structure of the data, the algorithm is not highly sensitive to the parameter setting. This means that the parameter choice is not overly strict. For example, when constructing $G_{\text{input}}$, we can choose k=50, and when constructing $G_{\text{output}}$, we can select k=10, achieving very good results. It is important to note that the k-NN graph is just one common method to capture the underlying structure of the data; it is neither the only nor the necessary approach. Other more advanced graph learning methods can also be used.

\section{Conclusion}\label{sect:conclusions}
In this paper, we view the noisy edges in graphs used from the perspective of adversarial attack. Building upon this viewpoint, we proposed a method based on robust nodes to mitigate the impact of noisy edges, grounded in adversarial robustness evaluation. Experimental results from real-world datasets show that our approach significantly boosts the performance of graph-based algorithms and outperforms the state-of-the-art methods in addressing noisy edges by a large margin.

%
%
%

\begin{thebibliography}{8}




\bibitem{von2007tutorial}
U.~Von~Luxburg, ``A tutorial on spectral clustering,'' \emph{Statistics and
  computing}, vol.~17, pp. 395--416, 2007.

\bibitem{premachandran2013consensus}
V.~Premachandran and R.~Kakarala, ``Consensus of k-nns for robust neighborhood
  selection on graph-based manifolds,'' in \emph{Proceedings of the IEEE
  Conference on Computer Vision and Pattern Recognition}, 2013, pp. 1594--1601.

\bibitem{roweis2000nonlinear}
S.~T. Roweis and L.~K. Saul, ``Nonlinear dimensionality reduction by locally
  linear embedding,'' \emph{science}, vol. 290, no. 5500, pp. 2323--2326, 2000.


\bibitem{Wang_2022_BMVC}
Y.~Wang and Z.~Feng, ``Towards scalable spectral clustering via
  spectrum-preserving sparsification,'' in \emph{33rd British Machine Vision
  Conference 2022, {BMVC} 2022, London, UK, November 21-24, 2022}.\hskip 1em
  plus 0.5em minus 0.4em\relax {BMVA} Press, 2022. 



\bibitem{yang2012affinity}
X.~Yang, L.~Prasad, and L.~J. Latecki, ``Affinity learning with diffusion on
  tensor product graph,'' \emph{IEEE transactions on pattern analysis and
  machine intelligence}, vol.~35, no.~1, pp. 28--38, 2012.

\bibitem{sarwar2001item}
B.~Sarwar, G.~Karypis, J.~Konstan, and J.~Riedl, ``Item-based collaborative
  filtering recommendation algorithms,'' in \emph{Proceedings of the 10th
  international conference on World Wide Web}, 2001, pp. 285--295.






\bibitem{goodfellow2018making}
I.~Goodfellow, P.~McDaniel, and N.~Papernot, ``Making machine learning robust
  against adversarial inputs,'' \emph{Communications of the ACM}, vol.~61,
  no.~7, pp. 56--66, 2018.

\bibitem{fawzi2018analysis}
A.~Fawzi, O.~Fawzi, and P.~Frossard, ``Analysis of classifiers’ robustness to
  adversarial perturbations,'' \emph{Machine learning}, vol. 107, no.~3, pp.
  481--508, 2018.

\bibitem{szegedy2013intriguing}
C.~Szegedy, W.~Zaremba, I.~Sutskever, J.~Bruna, D.~Erhan, I.~Goodfellow, and
  R.~Fergus, ``Intriguing properties of neural networks,'' \emph{arXiv preprint
  arXiv:1312.6199}, 2013.

\bibitem{nguyen2015deep}
A.~Nguyen, J.~Yosinski, and J.~Clune, ``Deep neural networks are easily fooled:
  High confidence predictions for unrecognizable images,'' in \emph{Proceedings
  of the IEEE conference on computer vision and pattern recognition}, 2015, pp.
  427--436.

\bibitem{moosavi2016deepfool}
S.-M. Moosavi-Dezfooli, A.~Fawzi, and P.~Frossard, ``Deepfool: a simple and
  accurate method to fool deep neural networks,'' in \emph{Proceedings of the
  IEEE conference on computer vision and pattern recognition}, 2016, pp.
  2574--2582.

\bibitem{feinman2017detecting}
R.~Feinman, R.~R. Curtin, S.~Shintre, and A.~B. Gardner, ``Detecting
  adversarial samples from artifacts,'' \emph{arXiv preprint arXiv:1703.00410},
  2017.

\bibitem{metzen2017detecting}
J.~H. Metzen, T.~Genewein, V.~Fischer, and B.~Bischoff, ``On detecting
  adversarial perturbations,'' \emph{arXiv preprint arXiv:1702.04267}, 2017.

\bibitem{yang2020ml}
P.~Yang, J.~Chen, C.-J. Hsieh, J.-L. Wang, and M.~Jordan, ``Ml-loo: Detecting
  adversarial examples with feature attribution,'' in \emph{Proceedings of the
  AAAI Conference on Artificial Intelligence}, vol.~34, no.~04, 2020, pp.
  6639--6647.

\bibitem{agarwal2019improving}
C.~Agarwal, A.~Nguyen, and D.~Schonfeld, ``Improving robustness to adversarial
  examples by encouraging discriminative features,'' in \emph{2019 IEEE
  International Conference on Image Processing (ICIP)}.\hskip 1em plus 0.5em
  minus 0.4em\relax IEEE, 2019, pp. 3801--3505.

\bibitem{liu2018adv}
X.~Liu, Y.~Li, C.~Wu, and C.-J. Hsieh, ``Adv-bnn: Improved adversarial defense
  through robust bayesian neural network,'' \emph{arXiv preprint
  arXiv:1810.01279}, 2018.



\bibitem{weng2018evaluating}
T.-W. Weng, H.~Zhang, P.-Y. Chen, J.~Yi, D.~Su, Y.~Gao, C.-J. Hsieh, and
  L.~Daniel, ``Evaluating the robustness of neural networks: An extreme value
  theory approach,'' \emph{arXiv preprint arXiv:1801.10578}, 2018.



\bibitem{cheng2021spade}
W.~Cheng, C.~Deng, Z.~Zhao, Y.~Cai, Z.~Zhang, and Z.~Feng, ``Spade: A spectral
  method for black-box adversarial robustness evaluation,'' in
  \emph{International Conference on Machine Learning}.\hskip 1em plus 0.5em
  minus 0.4em\relax PMLR, 2021, pp. 1814--1824.


\bibitem{spielman2012spectral}
D.~Spielman, ``Spectral graph theory,'' \emph{Combinatorial scientific
  computing}, vol.~18, p.~18, 2012.


\bibitem{shi2000normalized}
J.~Shi and J.~Malik, ``Normalized cuts and image segmentation,'' \emph{IEEE
  Transactions on pattern analysis and machine intelligence}, vol.~22, no.~8,
  pp. 888--905, 2000.

\bibitem{ng2001spectral}
A.~Ng, M.~Jordan, and Y.~Weiss, ``On spectral clustering: Analysis and an
  algorithm,'' \emph{Advances in neural information processing systems},
  vol.~14, 2001.

\bibitem{ben2001support}
A.~Ben-Hur, D.~Horn, H.~T. Siegelmann, and V.~Vapnik, ``Support vector
  clustering,'' \emph{Journal of machine learning research}, vol.~2, no. Dec,
  pp. 125--137, 2001.


  
\bibitem{papadimitriou1998combinatorial}
C.~H. Papadimitriou and K.~Steiglitz, \emph{Combinatorial optimization:
  algorithms and complexity}.\hskip 1em plus 0.5em minus 0.4em\relax Courier
  Corporation, 1998.

\bibitem{strehl2002cluster}
A.~Strehl and J.~Ghosh, ``Cluster ensembles---a knowledge reuse framework for
  combining multiple partitions,'' \emph{Journal of machine learning research},
  vol.~3, no. Dec, pp. 583--617, 2002.


\bibitem{szlam2009total}
A.~Szlam and X.~Bresson, ``A total variation-based graph clustering algorithm
  for cheeger ratio cuts,'' \emph{UCLA Cam Report}, pp. 09--68, 2009.


\bibitem{malkov2018efficient}
Y.~A. Malkov and D.~A. Yashunin, ``Efficient and robust approximate nearest
  neighbor search using hierarchical navigable small world graphs,'' \emph{IEEE
  transactions on pattern analysis and machine intelligence}, vol.~42, no.~4,
  pp. 824--836, 2018.


\bibitem{pakhira2014linear}
M.~K. Pakhira, ``A linear time-complexity k-means algorithm using cluster
  shifting,'' in \emph{2014 international conference on computational
  intelligence and communication networks}.\hskip 1em plus 0.5em minus
  0.4em\relax IEEE, 2014, pp. 1047--1051.

\bibitem{kyng2016approximate}
R.~Kyng and S.~Sachdeva, ``Approximate gaussian elimination for
  laplacians-fast, sparse, and simple,'' in \emph{2016 IEEE 57th Annual
  Symposium on Foundations of Computer Science (FOCS)}.\hskip 1em plus 0.5em
  minus 0.4em\relax IEEE, 2016, pp. 573--582.




\end{thebibliography}
%

\end{document}